\documentclass[conference]{IEEEtran}
\usepackage[utf8]{inputenc}
\usepackage{environ}
\usepackage{amsmath}
\usepackage{amssymb}
\usepackage{graphicx}
\usepackage{placeins}
\usepackage[table,xcdraw,usenames,dvipsnames]{xcolor}
\usepackage{dsfont}

\usepackage{hyperref}

\definecolor{source}{HTML}{154360}
\definecolor{target}{HTML}{17A589}

\usepackage{makecell}

\newcommand\numberthis{\addtocounter{equation}{1}\tag{\theequation}}

\NewEnviron{myequation*}{%
    \begin{equation*}
    \scalebox{1.2}{$\BODY$}
    \end{equation*}
    }

\NewEnviron{myequation}{%
\begin{equation}
\scalebox{1.2}{$\BODY$}
\end{equation}
}

\NewEnviron{myalign}{%
\begin{align*}
\scalebox{1.2}{\BODY}
\end{align*}
}

\ifCLASSINFOpdf
\else
\fi
\usepackage{algorithm, algpseudocode}
\algrenewcommand\algorithmicrequire{\textbf{Input:}}
\algrenewcommand\algorithmicensure{\textbf{Output:}}

\usepackage{stfloats}
\begin{document}
%
\title{Leveraging Data Geometry\\ to Mitigate CSM in Steganalysis}

\author{\IEEEauthorblockN{Rony Abecidan}
\IEEEauthorblockA{Univ. Lille, CNRS, Centrale Lille, \\ UMR 9189 CRIStAL,\\ F-59000 Lille, France\\
Email: rony.abecidan@univ-lille.fr}

\and

\IEEEauthorblockN{Vincent Itier}
\IEEEauthorblockA{IMT Nord Europe, Institut Mines-Télécom,\\ Centre for Digital Systems, Univ. Lille, CNRS, Centrale Lille,\\ UMR 9189 CRIStAL, F-59000 Lille, France\\
Email: vincent.itier@imt-nord-europe.fr}

\linebreakand
\and
\IEEEauthorblockN{Jérémie Boulanger}
\IEEEauthorblockA{Univ. Lille, CNRS, Centrale Lille, \\ UMR 9189 CRIStAL,\\ F-59000 Lille, France\\
Email:  jeremie.boulanger@univ-lille.fr}

\and
\IEEEauthorblockN{Patrick Bas}
\IEEEauthorblockA{Univ. Lille, CNRS, Centrale Lille, \\ UMR 9189 CRIStAL,\\ F-59000 Lille, France\\
Email:  patrick.bas@cnrs.fr}
\and
\IEEEauthorblockN{Tomáš Pevný}
\IEEEauthorblockA{Department of Computers and Engineering\\ Czech Technical University\\ Prague, Czech Republic\\
Email: pevnak@protonmail.ch}
}

%



\makeatletter
\newcommand{\linebreakand}{%
  \end{@IEEEauthorhalign}
  \hfill\mbox{}\par
  \mbox{}\hfill\begin{@IEEEauthorhalign}
}
\makeatother

\maketitle

\begin{figure}[b]
  \vspace{-0.3cm}
  \parbox{\hsize}{\em
  WIFS`2023, December, 4-7, 2023, Nuremberg, Germany.
  XXX-X-XXXX-XXXX-X/XX/\$XX.00 \ \copyright 2021 IEEE.
  }\end{figure}

\newcommand\NPB[1]{{\color{green!70!black}{ \small { [PB: ~#1~]}}}}

\newcommand\NRA[1]{{\color{red!70!black}{ \small { [RA: ~#1~]}}}}

\newcommand\NVI[1]{{\color{blue!70!black}{ \small { [VI: ~#1~]}}}}

\newcommand\NJB[1]{{\color{magenta!70!black}{ \small { [JB: ~#1~]}}}}

\newcommand\QF[1]{\textsc{qf}#1}

\begin{abstract}
  In operational scenarios, steganographers use sets of covers from various sensors and processing pipelines that differ significantly from those used by researchers to train steganalysis models. This leads to an inevitable performance gap when dealing with out-of-distribution covers, commonly referred to as Cover Source Mismatch (CSM). 
  In this study, we consider the scenario where test images are processed using the same pipeline. However, knowledge regarding both the labels and the balance between cover and stego is missing. Our objective is to identify a training dataset that allows for maximum generalization to our target.
  By exploring a grid of processing pipelines fostering CSM, we discovered a geometrical metric based on the chordal distance between subspaces spanned by DCTr features, that exhibits high correlation with operational regret while being not affected by the cover-stego balance. Our contribution lies in the development of a strategy that enables the selection or derivation of customized training datasets, enhancing the overall generalization performance for a given target.
  Experimental validation highlights that our geometry-based optimization strategy outperforms traditional atomistic methods given reasonable assumptions. Additional resources are available at \href{https://github.com/RonyAbecidan/LeveragingGeometrytoMitigateCSM}{github.com/RonyAbecidan/LeveragingGeometrytoMitigateCSM}.
\end{abstract}



%
\IEEEpeerreviewmaketitle

\section{Introduction}
\label{sec:intro}

Cover-source mismatch, also known as CSM, is a widely recognized phenomenon in modern steganalysis
\cite{giboulot} \cite{csm_chaumont} \cite{csm_fridrich}.
Traditional steganalysis models are typically trained on controlled cover distributions obtained from standardized databases like BOSSBASE~\cite{bossbase} or ALASKABASE~\cite{alaskabase}. However, in practical steganalysis scenarios, pinpointing the exact distributions to which the target images belong is quite challenging. The cover distributions, often referred to as cover sources, exhibit significant diversity due to various factors inherent to the image acquisition process. These factors include 
the sensor quality
, device settings
, captured content characteristics
, post-processing techniques applied
, as well as the subsequent compression steps
. These various stages of image development encompass a range of transformations, each influenced by specific parameters that affect the statistical properties of the resulting covers to any potential steganographic embedding.

\subsection{Mitigating CSM: Prior Arts}

It is shown in \cite{giboulot} that the processing pipeline is the cornerstone of the CSM, this pipeline encompasses a series of transformations commonly employed for aesthetic enhancement and compression purposes. However, in the domain of steganalysis, these transformations have a deep impact on the distribution of noise while maintaining the semantic integrity of the content. Despite the effectiveness of machine learning techniques in steganalysis, they often exhibit a high degree of sensitivity to the unique characteristics of the analyzed signals, thus giving rise to CSM, which can be seen in machine learning as a generalization problem.

To mitigate this issue, several approaches have been developed in the literature. In an ideal scenario, steganalysts would have access to cover distributions close to the ones of the images to scrutinize, enabling the creation of a specialized set of classifiers trained on these specific sources. For example when the CSM is only due to the JPEG quantization matrix or the camera model, Kodovsk{\'y} {\it et al.}~\cite{csm_fridrich} suggest choosing the classifier trained on the {\it closest} distribution w.r.t. the test images. They also suggest performing steganalysis on each evaluation image separately using the classifier trained on cover source that is most similar to that particular image. Various measures of source similarity are explored in this paper, and the most effective approach involved using a simple $\ell^2$ norm to compare the centers of gravity of the sources. Another approach presented in the literature by Pasquet {\it et al.}~\cite{csm_chaumont} involves the extraction of $K$ clusters called {\it islets} from a mixture of different sources and the training of $K$ specific ensemble classifiers. 
In a similar vein, Giboulot \textit{et al.}~\cite{giboulot} uses a multilinear classifier to assign the specific detector to be used for each evaluation image. 

In our recent study~\cite{wifs2022}, we report important conclusions regarding cover-source mismatch for various development parameters. Firstly, we highlighted the fact that denoising, downsampling, and post-resize sharpening play a significant role in the occurrence of CSM. 
Secondly, we observed that certain cover sources enable to build detectors that generalize more effectively than others. 
However, to the best of our knowledge, there is currently no existing work that comprehensively explores the relevance of specific sources in relation to particular targets.

\subsection{Contributions}

We propose an in-depth exploration of a grid of processing pipelines, focusing on our three parameters of interest, \textit{i.e.} denoising, downsampling, and post-resize sharpening. Our aim motivation is to understand how we can derive relevant 
training sets for operational steganalysis. To ensure realism, we consider a scenario in which the distribution of the covers under scrutiny is unknown. We assume access to an unlabeled set of target images with an unknown proportion of cover or stego images. Then, our goal is two-fold: 
\begin{itemize}
  \item Learn how to identify the most relevant training database compared to some others in relation to a given target.
  \item Develop a general strategy to build an appropriate training database for this target.
\end{itemize}

Motivated by these considerations, this paper represents the first attempt to address the CSM issue by proposing a framework for automatically generating appropriate steganalysis databases. Our work proposes the following contributions: 

\begin{itemize}
  \item We demonstrate that the geometrical properties of the target distribution plays a crucial role in identifying irrelevant sources that do not generalize well on target distributions.
  \item We present a simple yet efficient optimization algorithm for constructing relevant sources for specific targets. This algorithm leverages a geometry-based metric, the chordal distance, that evaluates the orthogonality between a source and a target.
\end{itemize}
 

The structure of the paper is as follows:
Section~\ref{sec:formalization} presents the formalization of our objective and introduces the manifold hypothesis that is core of the subsequent analysis.
In Section~\ref{sec:manifold}, a series of experiments are conducted to gain a deeper understanding of the components that contribute to an appropriate generalization from the source to the target.
In Section~\ref{sec:strategy}, a fresh perspective is presented to devise an optimization scheme that yields a target-specific relevant source leveraging a geometrical metric with interesting properties. The effectiveness of this novel approach is compared to conventional methods documented in prior literature.
Lastly, Section~\ref{sec:conclusion} serves as the conclusion, summarizing the main findings and contributions of this research.




\section{Formalization}
\label{sec:formalization}
\subsection{Problem formulation}
In accordance with \cite{sepak}, we define a processing pipeline as a vector $\omega \in \Omega$ that encompasses all the parameters associated with the pipeline, such as the downsampling, the denoising coefficient, the JPEG quality factor, and more. In the context of steganalysis, we introduce an additional parameter $\gamma$ to represent steganographer choices, including the embedding strategy and payload. For this particular task, machine learning models, acting as detectors are commonly used:

\begin{align*}
    f(x \mid \theta_{\omega,\gamma}) : \ & \mathcal{X} \rightarrow \{cover,stego\}. \\
    & x \mapsto y
  \end{align*}

Here, $\theta_{\omega,\gamma} \in \Theta$ represents the learned parameters using covers derived from the pipeline $\omega$ and potentially embedded following $\gamma$.

To effectively evaluate the CSM, two significant metrics have been introduced in \cite{giboulot} and \cite{sepak}:

\begin{itemize}
    \item The Intrinsic Difficulty of a source, quantified as the probability of error $P_E$ obtained after training on images from this source and evaluating on images from that same source:

    \begin{myequation}
        \mathbb E_{(x,y) \sim P((x,y)| \omega,\gamma)} (f(x \mid \theta_{\omega,\gamma} ) \neq y).
    \end{myequation}

    \item The Regret between a source $s$ and a target $t$, denoted as $R_{s,t} \geq 0$, or alternatively referred to as Source Inconsistency in \cite{giboulot}, defined as the difference between the $P_E$ we obtain by 
    training on $s$ and evaluating on $t$ and the Intrinsic Difficulty of $t$:
    
    {\large
  \begin{align*}
    \mathbb E_{(x,y) \sim P((x,y)| \omega_t,\gamma)} (f(x \mid \theta_{\omega_s,\gamma} ) \neq y), \\ - \mathbb E_{(x,y) \sim P((x,y)| \omega_t,\gamma)} (f(x \mid \theta_{\omega_t,\gamma} ) \neq y).  \numberthis
  \end{align*}
  }

\end{itemize}
In this study, we assume that we have access to target samples coming from an unknown distribution. It is important to note that the {\it cover-stego balance}, {\it i.e.} the balance between the covers and the stegos is also unknown, \textit{i.e.} that the CSM has to be solved in an unsupervised setting. 

Ideally, we want to find a source $s$ that minimizes the regret with our target $t$. However, since computing the regret in this scenario is infeasible, the best we could aim is the use of a metric $m$ correlated to the regret, computable without labels, and invariant to the cover-stego balance. Such a metric can be used as a benchmark for identifying and even creating the most effective source for our target dataset. 

\subsection{The manifold assumption}

Traditionally in machine learning, specific features of interest are used to represent the images that are living in very high dimension. As an example for steganalysis, the famous feature extractor DCTr \cite{dctr} enables the extraction of 8000 relevant features for each image to steganalyze. However, dimension 8000 is still large and it can be assumed that the DCTr features representing images coming from the same source belong to a manifold of significantly lower dimensionality than 8000~\cite{manifold}. Intuitively, the manifold underlying the source should be invariant to the cover-stego balance given a reasonable payload that does not move too far the stegos from the covers. In what follows, we provide empirical support for the importance of estimating the manifold supporting the target. By using PCA, one can approximate it as a linear space. This estimation yields valuable geometrical information that elucidates the observed mismatch with particular sources. 




\section{Understanding Source Relevance}
\label{sec:manifold}
\subsection{Experimental protocol}

\begin{figure*}[!t]
    \centerline{
    \includegraphics[width=0.95\textwidth]{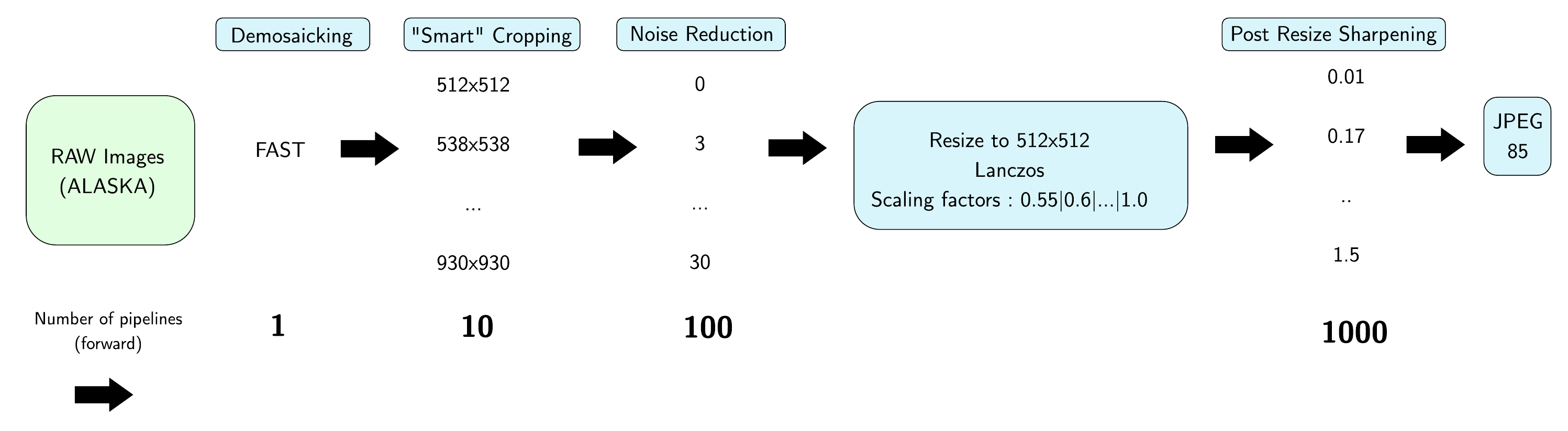}
    }
    \caption{Generation of $10^3$ pipelines. Note that the different scaling factors are only due to crops of different sizes.}
    \label{fig:pipelines}
  \end{figure*}

In \cite{wifs2022}, three operations promoting noise and content diversity were identified as important factors leading to CSM: denoising, downsampling and post-resize sharpening. In this study, we propose to investigate the factors that make a source more relevant than another for a specific target by creating a universe of sources obtained by developing RAW Images
, leveraging these three parameters of interest.

For the experiments presented here, we randomly extract 2000 RAW Images of various sizes from the ALASKA dataset~\cite{alaskabase}. Among these, 1000 RAW Images are dedicated to the source and the remaining are for the target. Out of the target images, 500 RAWs will be used to create {\it operational sets}, representing real-life scenarios where there are no pairs of covers and stegos, no labels, and no knowledge of the cover-stego balance. We will consider three balancing for this set: only covers, only stegos or an equal proportion of cover and stegos. The remaining 500 RAWs from the target will be used to generate samples dedicated to evaluation and to the computation of the regret, {\it i.e.} with equal numbers of cover and stego images.

The impact of JPEG compression on steganalysis is already well-documented in the literature \cite{giboulot}. In our experiments, we propose to apply a JPEG compression with a quality factor of 85 at the end of our processing pipelines. This choice facilitates the estimation of the manifolds underlying the sources, considering our limited sample size. We perform this final JPEG compression using Imagemagick\footnote{\href{https://imagemagick.org/}{imagemagick.org}} for full control over the compression process.

In total, we study $10^3$ combinations of parameters to explore a wide range of possibilities. It is important to note that we do not use a random cropping strategy as it often leads to low-textured covers in practice. Instead, we prefer the ``smart" cropping technique released in the ALASKA2 Challenge \cite{alaskabase}, which aims to select crops with highly textured areas. Details about the cover generation process are presented in Fig.~\ref{fig:pipelines}. 

Starting with the 1000 source RAWs, we create 1000 pairs of cover-stegos for training. The evaluation is performed using the 500 cover-stego pairs derived from the RAWs dedicated to evaluation. The  embedding is done using UERD \cite{uerd} with a payload of 0.5 bpnzac. To simplify the process and save computational resources, we then train linear classifiers using DCTR features \cite{dctr} from our sets of covers and stegos. This payload allows us to obtain cover sources with a reasonable Intrinsic Difficulty, ranging from 11\% to 30\%.

Once the cover distributions are generated, and the detectors are trained on each of them, we study CSM using a regret matrix $R$, where $R[s,t] = R_{s,t}$ and $s, t \in [|1,10^3|]^2$ indice all possible $10^3$ generated bases whereas there are seen as source or target. We then analyze the correlation between the regrets observed and some potential metrics we picked. Quantile plots help us visualize how the regret distribution evolves within a sliding window around different values of our metrics. We scale the metrics by dividing them by their maximum value to make them comparable. For each metric, we use points from 0 to 1 with a step of 0.01 and a window size of 0.3, which balances the tradeoff between localization and estimation accuracy.

We list below the three different metrics we used in this paper. The two first come from the litterature and the last one is new.

\subsection{$\ell^2$ between Centers of Gravity}

To evaluate if the source and the target are mismatching, \cite{csm_fridrich} recommends computing the $\ell^2$ norms between their centers of gravity, $\ell^2_{CG}$. This is a very appealing strategy, cheap to execute and easily explainable. However we expect its success to be highly dependent of the cover-stego balancing. For three operational target scenarios: only covers, only stegos, or a balanced proportion of both, we compute the $\ell^2_{CG}$ between the sources and operational targets in our universe. The $\ell^2_{CG}$ is comparable for the scenarios with one class, but it is changing significantly when we consider the two classes scenario. This makes sense as the center of gravity should not change much when we switch from using only covers to using only stegos as the operational target. However, a mix of cover and stego leads to a different gravity center than this common one.

\begin{figure}[h]
  \centerline{
  \includegraphics[width=1.2\columnwidth]{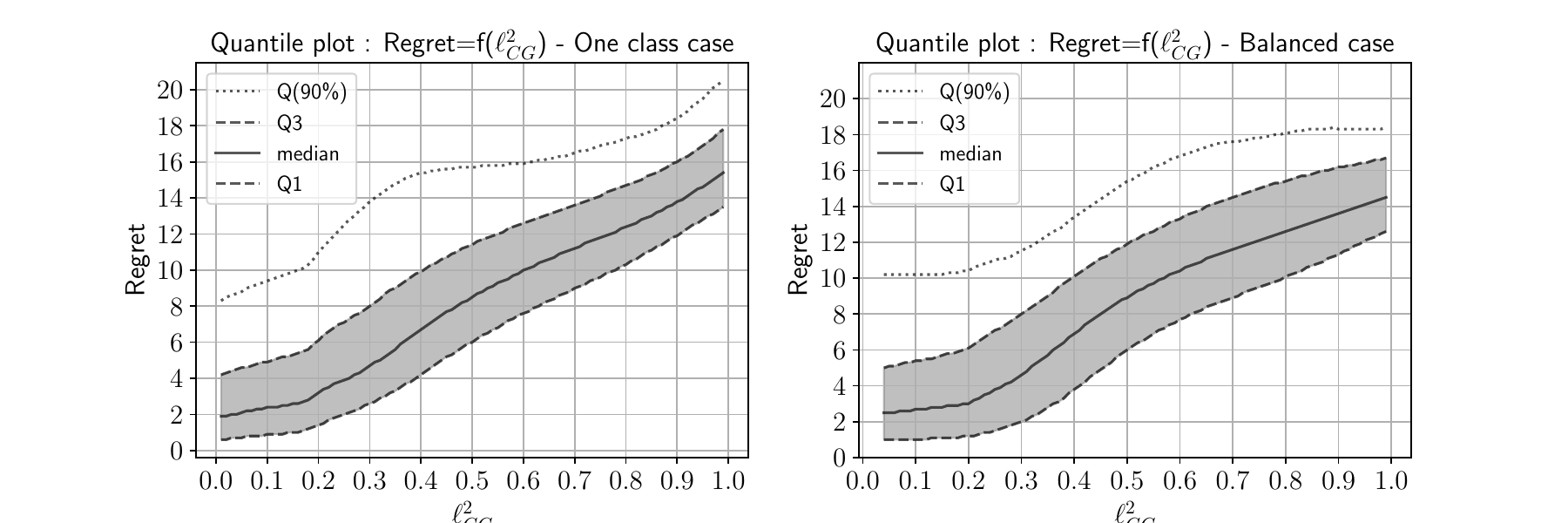}
    }
    \vspace*{-0.2cm}
  \caption{Quantile plots representing the evolution of the regret according to the $\ell^2_{CG}$ between sources and operational targets (dependent of the class-balancing). Q1 is the first quartile, Q3 is the third quartile and Q(90\%) is the 90th percentile.}
  \label{fig:L2_regret}
\end{figure}

We present in Fig.~\ref{fig:L2_regret} a quantile plot illustrating the evolution of the regrets in our universe according to the $\ell^2_{CG}$ between source and operational gravity centers. There is a clear correlation between the $\ell^2_{CG}$ and the regret. High $\ell^2_{CG}$ leads to high regrets while low $\ell^2_{CG}$ leads to low regrets. However, the regrets variance is non-negligeable and a low $\ell^2_{CG}$ does not ensure a low regret. 

\subsection{Maximum Mean Discrepancy}

As a second proxy to the regret, we propose considering a distance between distributions. One well-known metric in the literature is the maximum mean discrepancy (MMD). This metric is essentially an $L^2$ distance in a Hilbert space implicitly defined with a kernel function $k(x, y): \mathbb{R}^d \times \mathbb{R}^d \mapsto \mathbb{R}$ with real features of dimension $d$. The $\mathrm{MMD}$ is highly relevant in estimating the discrepancy between two distributions, but its effectiveness relies on its association with a characteristic kernel that complies with specific properties. When these properties are met, the metric is minimized if and only if the two distributions are identical. One compelling aspect of this metric is its simplicity and efficiency in estimation, requiring only a few samples from the distributions. For practitioners seeking a reliable and robust baseline, Feydy et al. \cite{feydy} recommend using the Energy Distance kernel: $$ k(x, y)=-\|x-y\|. $$
This well-known kernel in statistics produces appealing monotonic flows without the need for any parameter tuning. We can easily exploit it using the geomloss library \cite{geomloss}. We measure the $\mathrm{MMD}$ between each source-target pair in our universe, using again our three scenarios for the operational balancing. After the measurement, we find no significant difference of the $\mathrm{MMD}$ between each scenario. In Fig.~\ref{fig:MMD_regret}, we discover that the $\mathrm{MMD}$ is also well correlated to the regrets. However, the regret variance near the lowest values of the $\mathrm{MMD}$ is higher than the ones seen near the lowest values of the $\ell^2_{CG}$.  More clearly here, a low $\mathrm{MMD}$ does not imply a low regret.

\begin{figure}[h]
  \centerline{
  \includegraphics[width=1.0\columnwidth]{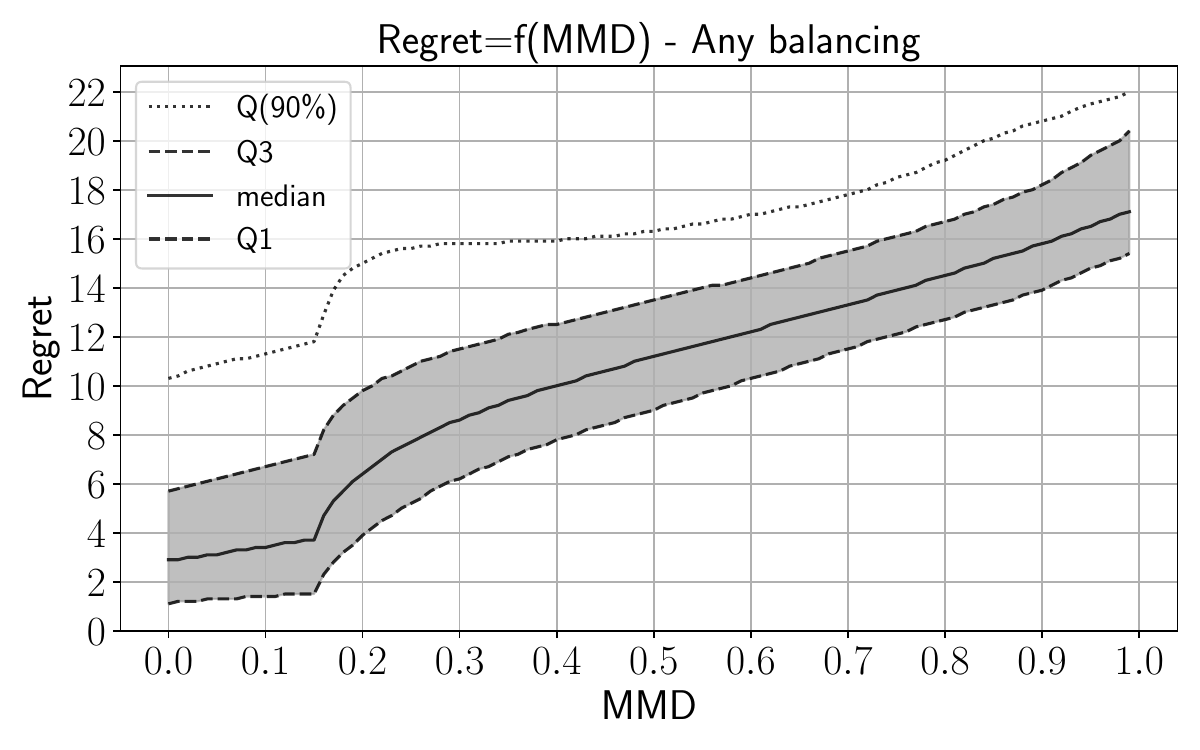}
    }
    \vspace*{-0.2cm}
  \caption{Quantile plots representing the evolution of the regret according to the $\mathrm{MMD}$ between sources and operational targets (dependent of the class-balancing). Q1 is the first quartile, Q3 is the third quartile and Q(90\%) is the 90th percentile.}
  \label{fig:MMD_regret}
\end{figure}


\begin{table*}[]
  {\scriptsize
  \begin{tabular}{c|cccccc||
  >{}c |ccccc||
  >{}c |ccccc|}
  \cline{2-19}
   &
    \multicolumn{6}{c||}{Only Covers} &
    \multicolumn{6}{c||}{Only Stegos} &
    \multicolumn{6}{c|}{Mix} \\ \cline{2-19}
   &
    \multicolumn{1}{c|}{Min} &
    \multicolumn{1}{c|}{Q1} &
    \multicolumn{1}{c|}{Q2} &
    \multicolumn{1}{c|}{Q3} &
    \multicolumn{1}{c|}{Max} &
    \%($R$\textgreater{}5) &
    \multicolumn{1}{c|}{Min} &
    \multicolumn{1}{c|}{Q1} &
    \multicolumn{1}{c|}{Q2} &
    \multicolumn{1}{c|}{Q3} &
    \multicolumn{1}{c|}{Max} &
    \%($R$\textgreater{}5) &
    \multicolumn{1}{c|}{Min} &
    \multicolumn{1}{c|}{Q1} &
    \multicolumn{1}{c|}{Q2} &
    \multicolumn{1}{c|}{Q3} &
    \multicolumn{1}{c|}{Max} &
   \%($R$\textgreater{}5) \\ \hline
  \multicolumn{1}{|c|}{Multi-classifier} &
    \multicolumn{1}{c|}{0.0} &
    \multicolumn{1}{c|}{0.4} &
    \multicolumn{1}{c|}{1.4} &
    \multicolumn{1}{c|}{2.8} &
    \multicolumn{1}{c|}{25} &
    8.7 &
     \multicolumn{1}{c|}{0.0} &
     \multicolumn{1}{c|}{0.4} &
     \multicolumn{1}{c|}{1.4} &
     \multicolumn{1}{c|}{2.8} &
     \multicolumn{1}{c|}{25} &
     8.7 &
      \multicolumn{1}{c|}{0.0} &
      \multicolumn{1}{c|}{0.4} &
      \multicolumn{1}{c|}{1.4} &
      \multicolumn{1}{c|}{2.8} &
      \multicolumn{1}{c|}{25} &
      8.7\\ \hline
      \multicolumn{1}{|c|}{Majority vote} &
    \multicolumn{1}{c|}{0.0} &
    \multicolumn{1}{c|}{6.6} &
    \multicolumn{1}{c|}{8.0} &
    \multicolumn{1}{c|}{9.4} &
    \multicolumn{1}{c|}{26.1} &
    \cellcolor[HTML]{FF81A5}89
     &
     \multicolumn{1}{c|}{0.0} &
     \multicolumn{1}{c|}{6.6} &
     \multicolumn{1}{c|}{8.0} &
     \multicolumn{1}{c|}{9.4} &
     \multicolumn{1}{c|}{26.1} &
     \cellcolor[HTML]{FF81A5}89
     &
     \multicolumn{1}{c|}{0.0} &
     \multicolumn{1}{c|}{6.6} &
     \multicolumn{1}{c|}{8.0} &
     \multicolumn{1}{c|}{9.4} &
     \multicolumn{1}{c|}{26.1} &
     \cellcolor[HTML]{FF81A5}89
     \\ \hline
  \multicolumn{1}{|c|}{min $\ell^2_{CG}$} &
    \multicolumn{1}{c|}{0.0} &
    \multicolumn{1}{c|}{1.1} &
    \multicolumn{1}{c|}{2.6} &
    \multicolumn{1}{c|}{4.0} &
    \multicolumn{1}{c|}{27} &
    16 &
    \multicolumn{1}{c|}{0.0} &
    \multicolumn{1}{c|}{1.0} &
    \multicolumn{1}{c|}{2.5} &
    \multicolumn{1}{c|}{4.0} &
    \multicolumn{1}{c|}{27} &
    14 &
    \multicolumn{1}{c|}{0.0} &
    \multicolumn{1}{c|}{0.8} &
    \multicolumn{1}{c|}{1.7} &
    \multicolumn{1}{c|}{3.5} &
    \multicolumn{1}{c|}{26} &
    20 \\ \hline
  \multicolumn{1}{|c|}{min MMD} &
    \multicolumn{1}{c|}{0.0} &
    \multicolumn{1}{c|}{0.5} &
    \multicolumn{1}{c|}{1.5} &
    \multicolumn{1}{c|}{3.1} &
    \multicolumn{1}{c|}{24} &
    10.2 &
    \multicolumn{1}{c|}{0.0} &
    \multicolumn{1}{c|}{0.3} &
    \multicolumn{1}{c|}{1.4} &
    \multicolumn{1}{c|}{2.7} &
    \multicolumn{1}{c|}{19} &
    7.9 &
    \multicolumn{1}{c|}{0.0} &
    \multicolumn{1}{c|}{0.4} &
    \multicolumn{1}{c|}{1.5} &
    \multicolumn{1}{c|}{2.9} &
    \multicolumn{1}{c|}{19} &
    8.7 \\ \hline
  \multicolumn{1}{|c|}{min $\mathrm{NSCD}$} &
    \multicolumn{1}{c|}{0.0} &
    \multicolumn{1}{c|}{0.1} &
    \multicolumn{1}{c|}{0.7} &
    \multicolumn{1}{c|}{1.5} &
    \multicolumn{1}{c|}{16} &
    \cellcolor[HTML]{8FF1B1}1.6 &
    \multicolumn{1}{c|}{0.0} &
    \multicolumn{1}{c|}{0.1} &
    \multicolumn{1}{c|}{0.7} &
    \multicolumn{1}{c|}{1.5} &
    \multicolumn{1}{c|}{16} &
    \cellcolor[HTML]{8FF1B1}1.5 &
    \multicolumn{1}{c|}{0.0} &
    \multicolumn{1}{c|}{0.1} &
    \multicolumn{1}{c|}{0.7} &
    \multicolumn{1}{c|}{1.4} &
    \multicolumn{1}{c|}{16} &
    \cellcolor[HTML]{8FF1B1}2.0 \\ \hline
  \end{tabular}
  }
  \vspace*{0.3cm}
\caption{Information about regret distributions obtained after applying different atomistic strategies in our universe of $10^3$ pipelines. $\%(R>5)$ is the ratio of regrets higher than $5 \%$ computed for all targets for each strategy (all results are in \%)}
  \label{tab:comparison}
  \end{table*}

\subsection{Assessing orthogonality between sources}

As the detector relies on the manifold supported by the source, any significant difference between the target and source manifolds can lead to a noticeable regret. This is the case when the two manifolds are not aligned.

%
%
%

Using the minimal angles between two orthogonal bases, namely principal angles, one can derive several metrics assessing misalignment between the two subspaces they span~\cite{chordal}. Considering two subspaces of different dimension $N$ and $K$, and the principal $\min(N, K)$ angles $\theta_i$ between those planes, we can then define the normalized squared chordal distance $\mathrm{NSCD}$~\cite{ye2016schubert}:

$$
\mathrm{NSCD}=\frac{\sum\limits_{i=1}^{\min(N,K)} \sin ^2\left(\theta_i\right)}{\min(N,K)}.
$$

\begin{figure}[b]
  \centerline{
  \includegraphics[width=1.0\columnwidth]{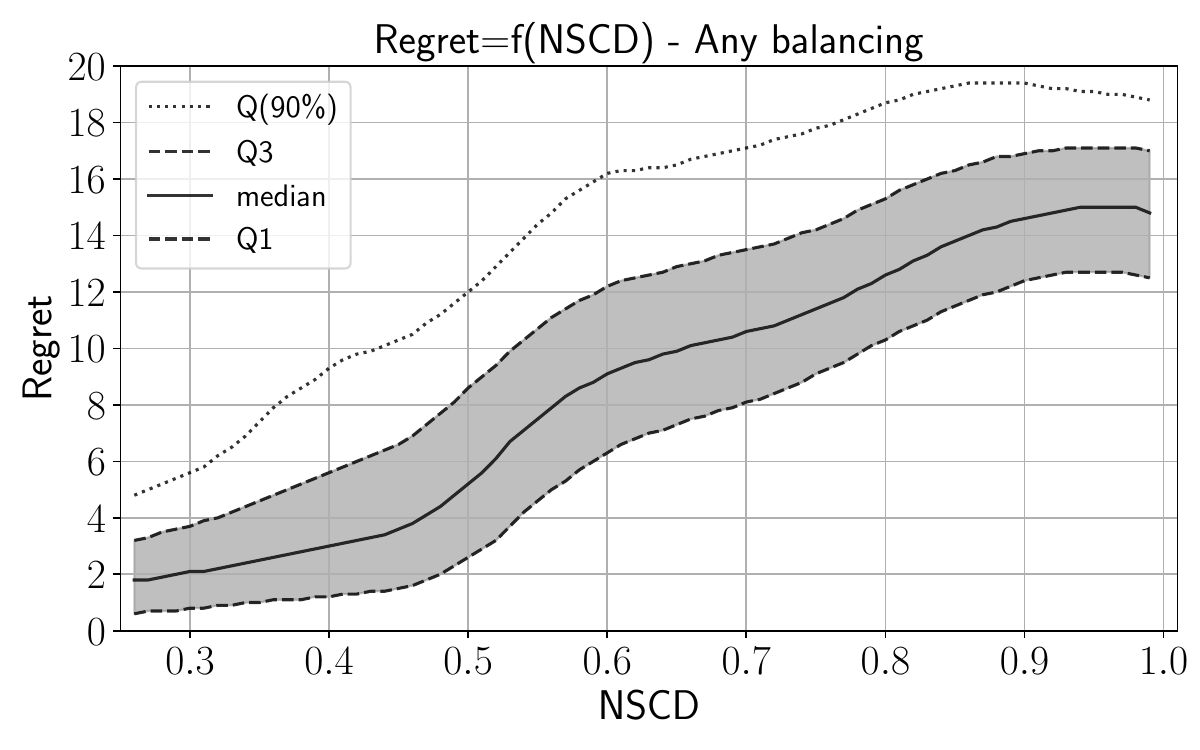}
    }
    \vspace*{-0.2cm}
  \caption{Quantile plots representing the evolution of the regret according to $\mathrm{NSCD}$ (invariant by balancing). Q1 is the first quartile, Q3 is the third quartile and Q(90\%) is the 90th percentile.}
  \label{fig:NSCD_regret}
\end{figure}

\noindent
When $\mathrm{NSCD}=0$ the two subspaces are equal while, when $\mathrm{NSCD}=1$, the two subspaces are orthogonal. 

For our experiments, we estimate the manifolds as the linear subspaces spanned by the orthogonal eigenvectors enabling to preserve 99.9\% of the variance computed through PCA (thus the dimension of subspace is not fixed). It is worth noting that this decision involves a careful tradeoff between dimensionality reduction and capturing sufficient variance to explain operational regrets. Afterwards, we compute the $\mathrm{NSCD}$ between every combination of source and target within our universe, taking again into account our three scenarios for the cover-stego balance. As one could expect, there is no significant difference of the $\mathrm{NSCD}$ w.r.t. the balance since the manifolds are not strongly modified after embedding. We observe in Fig.~\ref{fig:NSCD_regret} that there is a strong correlation between the~$\mathrm{NSCD}$ and the operational regret in our universe. This time, the regret variance around the lowest values of the~$\mathrm{NSCD}$ is pretty low. It appears to be the best proxy for the regret and clarifies well why a source is more suitable than other for a given target.


\section{Mitigating CSM with the chordal distance}
\label{sec:strategy}

\subsection{Comparison with classical atomistic strategies}

We propose here to compare the operational regrets obtained for each target of our universe given different strategies : 

\begin{itemize}

    \item Using a relevant source for each target image with the help of a multi-classifier trained to recognize the most representatives sources of our universe.
    \item Applying a majority vote to choose one relevant source for the target among the most representatives ones.
    \item Selecting the source that minimizes the $\ell^2_{CG}$ distance w.r.t. the operational set.
    \item Selecting the source that minimizes the $\mathrm{MMD}$ w.r.t. the operational set.
    \item Selecting the source that minimizes the $\mathrm{NSCD}$ w.r.t. the operational set.

\end{itemize}

Instead of using $10^3$ sources for the majority vote and the multi-classifier, which would severely degrade the performances due to a high number of candidates, we use our methodology from~\cite{wifs2022} to extract a small subset of representatives sources that can achieve less than 5\% regret for all targets. In practice, this enables to work on only 8 
sources in terms of generalization. Focusing on only these sources, the efficiency of these strategies is preserved while minimizing space and computation demands. 
Moreover, because there exist sources that are very close to each other, we want to evaluate our metric-based methods only on pairs of source/target which are enough different from each other. 
To address the bias induced by evalution on too close pairs
we remove candidates 
if their $\mathrm{MMD}$ is below the 1st decile $Q(10\%)$ of the $\mathrm{MMD}$ distribution that is invariant to operational balancing. This is done for the 3 last strategies to make a fair comparison. The two first strategies use representatives to select sources and are immune to this bias. 
Moreover, in our evaluations the sources will always have an equal proportion of cover and stego meanwhile the operational set will present different balancing.

We present, in Table~\ref{tab:comparison}, the statistical information relative to the distribution of regrets obtained following the different strategies.
Concerning the two first strategies, the majority vote is less satisfying compared to the others meanwhile the multi-classifier is among the best in terms of efficiency and invariance to the operational balancing. However, this particular strategy cruelly lacks of interpretability. 

On the other side, due to the invariance to cover stego balancing from the $\mathrm{MMD}$ and the $\mathrm{NSCD}$, it is not surprising to see very similar regret distributions in the 3 cases. The strategy using the $\ell^2$-norm is simple and already reasonable in many of the cases. However, its effectiveness depends on the cover stego balance. In the case where the operational target is made up of an equal proportion of covers and stegos, there is about 20\% of regrets observed higher than 5\% using this strategy whereas this lowers to 9\% with the $\mathrm{MMD}$ and 2\% with the $\mathrm{NSCD}$. It is worth mentioning that theoretically, there are situations where the $\mathrm{NSCD}$ is very low and the regret is high. One such example is when the target stems from a significant shift in the source manifold. Curiously, such cases are not present in our experiments. 

\subsection{Geometry-based optimization}

Since the $\mathrm{NSCD}$ exhibits a high correlation with the regret, we suggest leveraging it as a valuable tool for data-adaptation. The methodology involves starting with a set of RAW images, performing image development operations, computing the $\mathrm{NSCD}$ in relation to the target, and repeating the development to minimize the $\mathrm{NSCD}$ as much as possible. To optimize the selection of development parameters, we propose employing well-established heuristics such as simulated annealing. By testing this straightforward concept in our universe, specifically focusing on the most challenging scenarios where regrets are higher than 20\%, we observe satisfying results presented in Table~\ref{tab:geomoptim}. Out of the 30,000 cases studied, 88\% of the regrets obtained at the end of the optimization are below 5\%, with a maximum iteration count of 100. These experiments clearly indicate the superiority of this strategy over the previous method, which required the creation of $10^3$ sources. 

\begin{table}[h!]
    \begin{tabular}{|c|c|c|}
    \hline
    \textbf{Number of scenarios} & \textbf{\%($R_{final}\leq 5$)} & \textbf{Maximum iteration count} \\ \hline
    30.000                                                & 88\%                                      & 100                                    \\ \hline
    \end{tabular}
    \caption{Regret distribution outcomes of geometry-based optimization.}
    \label{tab:geomoptim}
\end{table}

\vspace*{-0.6cm}

\subsection{Impact of the target size}
With a very limited amount of target samples, the $\mathrm{NSCD}$ is not well estimated and leads to wrong source selections as shown in Table \ref{tab:nbsamples}. For instance, we observe that $\mathrm{NSCD}$ is not anymore invariant to operational balancing and much less useful when the amount of target samples is equal to 10. Speaking more generally, we observed through several experiments that when the noise level in the target rises, approximating the manifold becomes more challenging and a larger number of samples is required to achieve accurate estimation of the $\mathrm{NSCD}$. Here, we observe still an acceptable performance with 100 samples (with only $4.6\%$ regrets higher than 5\%). 

    \begin{table}[h!]
        \begin{tabular}{|c|cc|cc|cc|}
        \hline
                               & \multicolumn{2}{c|}{\textbf{Only Covers}}                                           & \multicolumn{2}{c|}{\textbf{Only Stegos}}                                           & \multicolumn{2}{c|}{\textbf{Mix}}                                                   \\ \hline
        \textbf{$N_{samples}$} & \multicolumn{1}{c|}{\textit{\textbf{Max}}} & \textit{\textbf{\%(R\textgreater{}5)}} & \multicolumn{1}{c|}{\textit{\textbf{Max}}} & \textit{\textbf{\%(R\textgreater{}5)}} & \multicolumn{1}{c|}{\textit{\textbf{Max}}} & \textit{\textbf{\%(R\textgreater{}5)}} \\ \hline
        10                     & \multicolumn{1}{c|}{27}                    & 45\%                                   & \multicolumn{1}{c|}{27}                    & 44\%                                   & \multicolumn{1}{c|}{24}                    & 23.8\%                                 \\ \hline
        100                    & \multicolumn{1}{c|}{16}                    & 2.7\%                                  & \multicolumn{1}{c|}{16}                    & 2.5\%                                  & \multicolumn{1}{c|}{16}                    & 4.6\%                                  \\ \hline
        500                    & \multicolumn{1}{c|}{16}                    & 1.6\%                                  & \multicolumn{1}{c|}{16}                    & 1.5\%                                  & \multicolumn{1}{c|}{16}                    & 2\%                                    \\ \hline
        \end{tabular}
        \caption{Information about the regret distributions obtained selecting the source minimizing the $\mathrm{NSCD}$ according to the number of operational samples available. The column max is for the maximum regret observed according to each case.}
        \label{tab:nbsamples}
        \end{table}


\vspace*{-0.5cm}
\section{Conclusions and Perspectives}
\label{sec:conclusion}

In this paper, we explore a universe of sources derived from processing pipelines, focusing on denoising, downsampling, and post-resize sharpening parameters, with the primary aim of understanding what makes some sources more relevant than others in operational steganalysis. In our scenario, the cover distribution of target images is unknown and the steganalyst is unaware of the proportion of covers and stegos. The main goal is to develop a general strategy for designing training databases tailored to specific targets which may have different distributions but allow us to minimize the regret. We show that the geometrical properties of the target play a key role in identifying relevant and irrelevant sources that respectively do or do not help to generalize well on target distributions. To this end, we present the $\mathrm{NSCD}$, a metric assessing the orthogonality between subspace spanned by source and target DCTr features. 
Our optimization algorithm, which uses the $\mathrm{NSCD}$ proved to be efficient in guiding source selection.


Looking ahead, we want to explore whether the $\mathrm{NSCD}$ can be used to generalize on images from entirely unknown distributions, such as those obtained from platforms like Flickr. Moreover, combining the metrics exhibited here to create a promising new composite metric could yield even more powerful results. At last, our ongoing research investigates whether $\mathrm{NSCD}$ can help to identify optimal sources with state-of-the-art deep neural networks for steganalysis. Encouraging results support further investigation.
In conclusion, this paper serves as an initial step toward database adaptation for steganalysis, i.e. generating automatically tailored steganalysis sources for operational targets. 



\section*{Acknowledgments}

Our experiments were possible thanks to computing means of IDRIS through the resource allocation 2022-AD011013285R1 assigned by GENCI. This work received funding from the European Union’s
Horizon 2020 research and innovation program under grant agreement No 101021687 (project “UNCOVER”) 
and the French Defense \& Innovation Agency. The work of Tomas Pevny was supported by Czech Ministry of Education 19-29680L.





\bibliographystyle{IEEEtran}


\bibliography{refs}

\end{document}